\documentclass[sigconf]{acmart}

\AtBeginDocument{%
  }

\copyrightyear{2024} 
\acmYear{2024} 
\setcopyright{rightsretained} 
\acmConference[MLCAD '24]{2024 ACM/IEEE International Symposium on Machine Learning for CAD}{September 9--11, 2024}{Salt Lake City, UT, USA}
\acmBooktitle{2024 ACM/IEEE International Symposium on Machine Learning for CAD (MLCAD '24), September 9--11, 2024, Salt Lake City, UT, USA}
\acmDOI{10.1145/3670474.3685966}
\acmISBN{979-8-4007-0699-8/24/09}
\begin{document}

\title{Chain-of-Descriptions: Improving Code LLMs for VHDL Code Generation and Summarization}

\author{Prashanth Vijayaraghavan}

\affiliation{%
  \institution{IBM Research}
  \city{San Jose}
  \state{CA}
  \country{USA}
}
\email{prashanthv@ibm.com}
\author{Apoorva Nitsure}
\affiliation{%
  \institution{IBM Research}
  \city{San Jose}
  \state{CA}
  \country{USA}
}
\email{Apoorva.Nitsure@ibm.com}

\author{Charles Mackin}
\affiliation{%
  \institution{IBM Research}
  \city{San Jose}
  \state{CA}
  \country{USA}
}
\email{charles.mackin@ibm.com}

\author{Luyao Shi}
\affiliation{
  \institution{IBM Research}
  \city{San Jose}
  \state{CA}
  \country{USA}
}
\email{luyao.shi@ibm.com}

\author{Stefano Ambrogio}
\affiliation{%
  \institution{IBM Research}
  \city{San Jose}
  \state{CA}
  \country{USA}}
\email{ambrogio.stefano@gmail.com}

\author{Arvind Haran}
\affiliation{%
  \institution{IBM}
  \city{Austin}
  \state{TX}
  \country{USA}}
\email{aharan@us.ibm.com}

\author{Viresh Paruthi}
\affiliation{%
  \institution{IBM}
  \city{Austin}
  \state{TX}
  \country{USA}}
\email{vparuthi@us.ibm.com}

\author{Ali Elzein}
\affiliation{%
  \institution{IBM}
  \city{Austin}
  \state{TX}
  \country{USA}}
\email{elzein@us.ibm.com}

\author{Dan Coops}
\affiliation{%
  \institution{IBM}
  \city{Austin}
  \state{TX}
  \country{USA}}
\email{coops@us.ibm.com}
\author{David Beymer}
\affiliation{%
 \institution{IBM Research}
  \city{San Jose}
  \state{CA}
  \country{USA}}
\email{beymer@us.ibm.com}

\author{Tyler Baldwin}
\affiliation{%
 \institution{IBM Research}
  \city{San Jose}
  \state{CA}
  \country{USA}}
\email{tbaldwin@us.ibm.com}

\author{Ehsan Degan}
\affiliation{%
  \institution{IBM Research}
  \city{San Jose}
  \state{CA}
  \country{USA}}
\email{edehgha@us.ibm.com}

\renewcommand{\shortauthors}{Vijayaraghavan et al.}


\begin{abstract}

Large Language Models (LLMs) have become widely used across diverse NLP tasks and domains, demonstrating their adaptability and effectiveness. In the realm of Electronic Design Automation (EDA), LLMs show promise for tasks like Register-Transfer Level (RTL) code generation and summarization. However, despite the proliferation of LLMs for general code-related tasks, there's a dearth of research focused on evaluating and refining these models for hardware description languages (HDLs), notably VHDL. In this study, we evaluate the performance of existing code LLMs for VHDL code generation and summarization using various metrics and two datasets -- VHDL-Eval and VHDL-Xform. The latter, an in-house dataset, aims to gauge LLMs' understanding of functionally equivalent code. Our findings reveal consistent underperformance of these models across different metrics, underscoring a significant gap in their suitability for this domain. To address this challenge, we propose Chain-of-Descriptions (CoDes), a novel approach to enhance the performance of LLMs for VHDL code generation and summarization tasks. CoDes involves generating a series of intermediate descriptive steps based on: (i) the problem statement for code generation, and (ii) the VHDL code for summarization. These steps are then integrated with the original input prompt (problem statement or code) and provided as input to the LLMs to generate the final output. Our experiments demonstrate that the CoDes approach significantly surpasses the standard prompting strategy across various metrics on both datasets. This method not only improves the quality of VHDL code generation and summarization but also serves as a framework for future research aimed at enhancing code LLMs for VHDL.

\end{abstract}



\maketitle
\section{Introduction}

The importance of Register-Transfer Level (RTL) designs in Electronic Design Automation (EDA) cannot be overstated. RTL designs form the backbone of digital circuit design, enabling the synthesis of high-level functional specifications into hardware implementations. Automating RTL code generation and summarization, particularly in VHDL, can significantly accelerate design synthesis and enhance understanding of functionality. Automated VHDL code generation can reduce the time and effort required for manual coding, while effective summarization can help designers quickly grasp the purpose and operation of complex circuits.

Large Language Models (LLMs) are increasingly popular across various domains, including natural language processing, image captioning, and lately, code generation and summarization. While code-specific LLMs excel in handling diverse programming tasks, there's a notable gap in research concerning VHDL, despite some models being trained on limited VHDL data. This lack of emphasis hinders the effectiveness of existing LLMs in addressing VHDL's unique challenges. The challenges with using code LLMs for VHDL are twofold. First, these models lack specific tuning for VHDL due to limited exposure to VHDL data and a lack of robust evaluation methods for this language. Second, they struggle to grasp functional equivalence, leading to discrepancies in code summarization, as they often prioritize syntactic similarities over functional equivalence. To overcome these challenges, we employ an evaluation dataset curated by \cite{pralav_lshi}, merging publicly available VHDL data with the VHDL-translated Verilog-Eval dataset. Additionally, we aggregate an in-house VHDL-Xform dataset, comprising VHDL codes and diverse code clones, aiming to assess LLMs' comprehension of functionally equivalent code and their ability to generate similar summaries. We conduct a comprehensive evaluation across various LLMs, ranging from Code-specific LLMs to Chat-based models for code generation and summarization tasks. Our evaluation metrics include not only the standard pass@k metric but also the self-consistency score, which measures the LLMs' capability to bidirectionally perform both generation and summarization. Furthermore, we measure the LLMs' performance on code sumarization using ROUGE-L and LLM Preference rate (PR) metrics. PR is computed with the help of a judge LLM to compare LLM-generated summaries with the reference and evaluate if summaries of functionally equivalent code clones capture the same functionality. 
Our comprehensive evaluation reveals unsatisfactory results from existing models, indicating substantial room for improvement. To address these challenges, we propose the Chain-of-Descriptions (CoDes) strategy. CoDes utilizes a series of intermediate descriptive steps based on the problem statement or VHDL code, depending on the task. These intermediate steps aid the model in better understanding the problem or code, leading to more accurate and coherent outputs for code generation and summarization tasks.

Our key contributions are summarized as follows: (a) Introduction of the VHDL-Xform dataset, which includes diverse code clones aimed at assessing LLMs' comprehension of functionally equivalent code. (b) Zero-shot evaluation of various LLMs using both the VHDL-Eval and VHDL-Xform datasets for VHDL code generation and summarization.
(c) Investigation of the Chain-of-Descriptions (CoDes) strategy to enhance LLMs' performance in VHDL code generation and summarization, thereby establishing a benchmark for future research in LLMs for VHDL.

\section{Related Work}

We categorize the literature into two main domains: Large Language Models (LLMs) for Code and chip design. LLMs for Code have made significant strides due to transformer-based deep neural networks \cite{vaswani2017attention}. Trained on extensive code bases alongside natural language, these models exhibit remarkable capabilities, discerning target languages from various prompts, including instructions and code snippets. Notable models like NVIDIA's MegatronLM \cite{shoeybi2019megatron}, StarCoder \cite{li2023starcoder} and Salesforce's CodeGen \cite{nijkamp2022codegen} are specifically tuned for coding tasks, with recent advancements including instruction-tuned and code-specific models like CodeLlama \cite{roziere2023code} and Granite \cite{mishra2024granite}. In line with this research, researchers explore integrating LLMs into chip design challenges, such as Dave's initiative \cite{pearce2020dave} in generating Verilog from English using GPT-2. Recent efforts \cite{thakur2023benchmarking,thakur2024verigen} demonstrate the effectiveness of fine-tuned open-source LLMs, like CodeGen \cite{nijkamp2022codegen}, for Verilog tasks. The Verilog-Eval benchmark \cite{liu2023verilogeval} illustrates potential improvements in Verilog code generation through supervised fine-tuning with LLM-generated synthetic problem-code pairs. Another line of work in the Chip design domain explored conversational approaches \cite{blocklove2023chip} for designing and verifying microprocessors, employing GPT-4 \cite{achiam2023gpt} and GPT-3.5. Despite GPT-4's ability to produce high-quality code, challenges in error understanding and fixing persist. ChipNemo \cite{liu2023chipnemo} demonstrates that domain-adaptive pretraining of language models can lead to superior performance in engineering assistant chatbots and EDA scripts generation. Despite the growing interest in leveraging LLMs for EDA, such explorations are limited for hardware languages like VHDL. This study evaluates various LLMs for VHDL code generation and summarization, introducing the Chain-of-Descriptions framework. Inspired by prior work on Chain-of-Thoughts \cite{wei2022chain,dhuliawala2023chain}, this framework investigates LLMs' ability to plan and execute VHDL code generation and summarization effectively, contributing to the limited body of research exploring VHDL-focused frameworks.

\section{Datasets}
\subsection{VHDL-Eval}
The VHDL-Eval dataset, as introduced in \cite{pralav_lshi}, was created by converting a set of Verilog evaluation problems into VHDL and compiling publicly accessible VHDL problems. This compilation resulted in a total of 202 problems. The dataset consists of meticulously curated self-verifying testbenches tailored for this amalgamated VHDL problem set. Additional details about the dataset specifications can be found in Appendix \ref{sec:app_vhdleval}.  

\subsection{VHDL-Xform}
In this work, we prepare a dataset that leverages publicly available Github VHDL repositories and filter them based on licensing information such that only those repositories that allow some form of free usage of code are used. As a next step, we apply different code transformation strategies based on the concept of code clones referring to a block of code that is identical or very similar to another block of code. Drawing from the prior literature on the categories of code clones \cite{bellon2007comparison,saini2018oreo} ranging from purely textual (Type-1) to
purely semantic (Type-4), we apply Type-2, Type-3 and Type-4 transformations to the VHDL code, as defined below. 
Appendix \ref{sec:app_clone} includes sample transformations for each category of code clones, along with the corresponding data statistics. Given a particular VHDL code, we apply the following transformations:

\noindent \textbf{Type-2:} Focuses on modifying identifiers and literals in VHDL code by parsing it and adjusting the entity, architecture, and port names consistently across the code. Unique identifiers are extracted using a VHDL parser, including entity, package, process, architecture, and port names, and used for naming suggestions by an LLM. Single-character identifiers are replaced by those suggested by the LLM, maintaining developer conventions. For snake or camel case identifiers, options include permuting, abbreviating, or using LLM suggestions to preserve program behavior.

\noindent \textbf{Type-3:} Involves changing statements that vary at the statement level in code snippets. Functionally inert code is introduced, and port or signal name declarations are reordered. These modifications ensure code integrity without compromising compilation or functionality.

\noindent \textbf{Type-4:} Addresses semantically similar code snippets with minimal lexical overlap using a back-translation approach. VHDL is converted to Verilog and back using GHDL and ICARUS Iverilog tools, altering identifier names and introducing temporary states to capture semantic differences while minimizing lexical similarities.




\section{Zero-Shot Evaluation of Code LLMs}
\subsection{LLM Baselines}
We compare LLMs across 3 categories-- (a) \textbf{Code Models}: CodeLlama-34b, Granite-Code-34b, Deepseek-33b, and Granite-Code-20b; (b) \textbf{Natural Language (NL) Instruct Models}: Mixtral-8x7b, and Mistral-7b; and (c) \textbf{Chat Models}: Granite-Chat-13b, and Llama-2-Chat-13b. More details are provided in the Appendix \ref{sec:bl} and \ref{sec:eval_settings}.

\subsection{Metrics}
We explore different metrics for evaluating LLMs in code generation and summarization tasks.

\subsubsection{Pass@k}
We use the \textit{Pass@k} metric, a common measure in recent code generation studies \cite{chen2021evaluating}. This metric checks if any of the $k$ generated code samples are functionally correct. In this work, we primarily report the Pass@1 scores based on the assessment of functional correctness of the generated VHDL code using the following two methods:

\textit{Self-Verifying Testbenches (TB)}:
We use self-verifying testbenches from the VHDL-Eval dataset to ensure the generated VHDL code passes all test cases in their testbench files, verifying critical functionalities. However, these test cases might not be completely exhaustive due to limitations in covering all possible scenarios, edge cases, and input combinations. To facilitate this automated testing, we employ VUnit and GHDL, which provide a robust environment for executing the available test cases and ensuring the VHDL code meets the specified requirements under the tested conditions.

\textit{Sequential Equivalence Checking (SEC)}: We apply the Sequential Equivalence Checking (SEC) methodology \cite{baumgartner2006scalable} to verify the functional correctness of the model response in the code generation task. SEC is widely used in hardware development to perform a true sequential check of input/output equivalence between two designs. It involves using a coordinated set of algorithms on two designs, represented as netlists, to validate that they produce identical sequences of valuations at their primary outputs under all possible identical sequences of valuations at their primary inputs, starting from correlated initial states. In this metric, we consider the model response ``functionally correct'' if it is proven to be sequentially equivalent to the reference implementation, referred to as ``canonical\_solution'' in the VHDL-Eval dataset. Any counterexample traces produced indicate a divergence in functionality from the reference implementation.
\label{sec:sec_metric}

\subsubsection{Self-Consistency}
Based on a prior study \cite{min2023beyond}, we evaluate the code LLMs for VHDL based on a variant of IdentityChain approach. In our case, we start from a canonical solution $pl_0$ in the VHDL-Eval dataset, instruct the model to generate a natural language summary $nl_0$, and generate a new program $pl_1$ from the summary. If the test outputs of $pl_1$ do not match those of $pl_0$, then the model is not self-consistent. A trustworthy model should be self-consistent as it assesses the model's ability to bidirectionally translate between the two spaces. To evaluate the self-consistency of a model in this work, we set the length of the chain $n=1$ and leverage the canonical solutions from the VHDL-Eval dataset as the evaluation set. We index the inputs in the evaluation set by $j, 1 \leq j\leq m$, where $m$ is the size of the VHDL-Eval dataset. For an input $pl_{0,j}$ in the evaluation set, we check its corresponding semantic equalities $sem(pl_0) = sem(nl_0) = sem(pl_1)$ by using the sequential equivalence checking (SEC) mentioned in Section \ref{sec:sec_metric}. We use a binary output $sc_{1,j} \in \{0, 1\}$ to indicate the LLM is self-consistent w.r.t to $pl_{0,j}$. Formally, $SC_1 := \frac{\sum_{i=1}^m{sc_{1,i}}}{m}$. 
\subsubsection{ROUGE-L}
This metric measures the similarity between a generated code summary and a reference summary using the longest common subsequence (LCS). We report {ROUGE-L} scores ($R_L$) for the VHDL-Xform dataset, comparing the LLM-generated summary of the transformed code to the LLM-generated summary of the original code as the reference.

\subsubsection{LLM Preference Rate}
\label{sec:pr}
Lexical similarity metrics like ROUGE-L often fail to capture semantic similarity, leading to inconsistent evaluations. To address this, we use an indirect evaluation strategy with a powerful LLM as the judge, following the LLM preference rate method from \cite{yuan2023evaluating}. The judge LLM evaluates the generated code summary against the reference summary in the VHDL-Eval dataset and checks if the LLM-generated summary of the transformed code aligns with both the original code and its LLM-generated summary in the VHDL-Xform dataset. The preference rate (PR) is calculated as the frequency with which the LLM-generated summary is preferred, measuring the model's capability in code summarization across both datasets. In this work, we use Llama-3-70b-instruct\footnote{\url{https://huggingface.co/meta-llama/Meta-Llama-3-70B-Instruct}} as the judge LLM; more details are provided in Appendix \ref{sec:pr_more}.

   



\begin{table}[]
\centering
\small
\caption{Zero-Shot Evaluation Results for code generation \& summarization using VHDL-Eval and VHDL-Xform dataset. \label{tab:zs_eval}}
\begin{tabular}{@{}lccclcc@{}}
\toprule
\multicolumn{1}{c|}{\textbf{Models}}  & \multicolumn{4}{c|}{\textbf{VHDL-Eval}}                                                                                                                                                                                                         & \multicolumn{2}{c}{\textbf{VHDL-Xform}}                            \\ \midrule
\multicolumn{1}{l|}{}                 & \begin{tabular}[c]{@{}c@{}}Pass@1\\ (TB)\end{tabular} & \begin{tabular}[c]{@{}c@{}}Pass@1\\ (SEC)\end{tabular} & \begin{tabular}[c]{@{}c@{}}SC$_1$\\ (\%)\end{tabular} & \multicolumn{1}{l|}{\begin{tabular}[c]{@{}l@{}}PR\\ (\%)\end{tabular}} & R$_L$         & \begin{tabular}[c]{@{}c@{}}PR \\ (\%)\end{tabular} \\ \midrule
\multicolumn{7}{c}{\textbf{Code Models}}                                                                                                                                                                                                                                                                                                                     \\ \midrule
\multicolumn{1}{l|}{CodeLlama-34b}    & 0.186                                                 & 0.176                                                  & 20.6                                                  & \multicolumn{1}{l|}{\textbf{34.7}}                                     & 36.63         & \textbf{36.6}                                      \\
\multicolumn{1}{l|}{Granite-Code-34b} & \textbf{0.192}                                        & \textbf{0.187}                                         & \textbf{23.5}                                         & \multicolumn{1}{l|}{33.8}                                              & \textbf{38.6} & 35.7                                               \\
\multicolumn{1}{l|}{Deepseek-33b}     & 0.164                                                 & 0.145                                                  & 14.8                                                  & \multicolumn{1}{l|}{25.7}                                              & 35.1          & 28.6                                               \\
\multicolumn{1}{l|}{Granite-Code-20b} & 0.085                                                 & 0.072                                                  & 12.1                                                  & \multicolumn{1}{l|}{26.6}                                              & 31.7          & 28.1                                               \\ \midrule
\multicolumn{7}{c}{\textbf{Instruct Models}}                                                                                                                                                                                                                                                                                                                 \\ \midrule
\multicolumn{1}{l|}{Mixtral-8x7b}     & 0.056                                                 & 0.042                                                  & 8.3                                                   & \multicolumn{1}{c|}{12.9}                                              & 19.6          & 14.8                                               \\
\multicolumn{1}{l|}{Mistral-7b}       & 0.042                                                 & 0.031                                                  & 3.9                                                   & \multicolumn{1}{c|}{5.2}                                               & 18.2          & 6.6                                                \\ \midrule
\multicolumn{7}{c}{\textbf{Chat Models}}                                                                                                                                                                                                                                                                                                                     \\ \midrule
\multicolumn{1}{l|}{Granite-Chat-13b} & 0.061                                                 & 0.059                                                  & 7.5                                                   & \multicolumn{1}{l|}{20.0}                                              & 28.1          & 19.5                                               \\
\multicolumn{1}{l|}{Llama-2-Chat-13b} & 0.059                                                 & 0.050                                                  & 6.6                                                   & \multicolumn{1}{l|}{19.5}                                              & 26.6          & 18.5                                               \\ \bottomrule
\end{tabular}
\end{table}

\subsection{Zero-Shot Evaluation Results}
Table \ref{tab:zs_eval} displays the zero-shot evaluation outcomes of various large language models (LLMs) on code generation and summarization tasks using VHDL-Eval and VHDL-Xform datasets. Evaluation metrics encompass Pass@1 (TB and SEC), Self-Consistency score (SC$_1$), LLM Preference Rate (PR), and ROUGE-L score (R$_L$). Granite-Code-34b and CodeLlama-34b emerge as top performers, with Granite-Code-34b showing a slight edge, boasting approximately 3\% and 6\% higher Pass@1 (TB) and Pass@1 (SEC) rates, respectively. The difference in the Pass@1 metric between the TB and SEC methods arises from the non-exhaustive nature of the testbenches and the inherent rigor of the SEC approach. Granite-Code-34b achieves the highest self-consistency score (SC$_1$) at 23.5\% and a notable ROUGE-L score (R$_L$) of 38.6, despite CodeLlama-34b having a higher LLM Preference Rate (PR) of 34.7\%. Compared to Deepseek-33b, the smaller Granite-Code-20b model demonstrates comparable performance in VHDL code summarization, with a SC$_1$ of 12.1\% and PR of 26.6\%. Instruct models like Mixtral-8x7b and Mistral-7b excel in summarization tasks, reflecting their training focus on natural languages. Similarly, Chat models such as Granite-Chat-13b and Llama-2-Chat-13b show a similar trend, with Granite-Chat-13b slightly outperforming its counterpart. We also note that the $SC_1$ scores are higher than the Pass@1 (SEC) metric, despite both using the SEC method. This is because the descriptions generated by the LLMs from the code are more straightforward than the problem statements in the VHDL-Eval dataset, which require the LLMs to make inferences (refer Appendix \ref{sec:app_vhdleval}). Clear descriptions improve model performance, resulting in higher $SC_1$ scores. However, the performance of these models significantly lags behind their capabilities in more popular languages like Python. This underscores the necessity for specialized approaches to enhance LLMs for VHDL tasks. To address this, we propose a novel framework called Chain-of-Description (CoDes) aiming to improve LLM performance in VHDL code summarization and generation, potentially bridging the observed performance gap.

\begin{figure*}[!ht]
    \centering
    \includegraphics[width=0.75\textwidth]{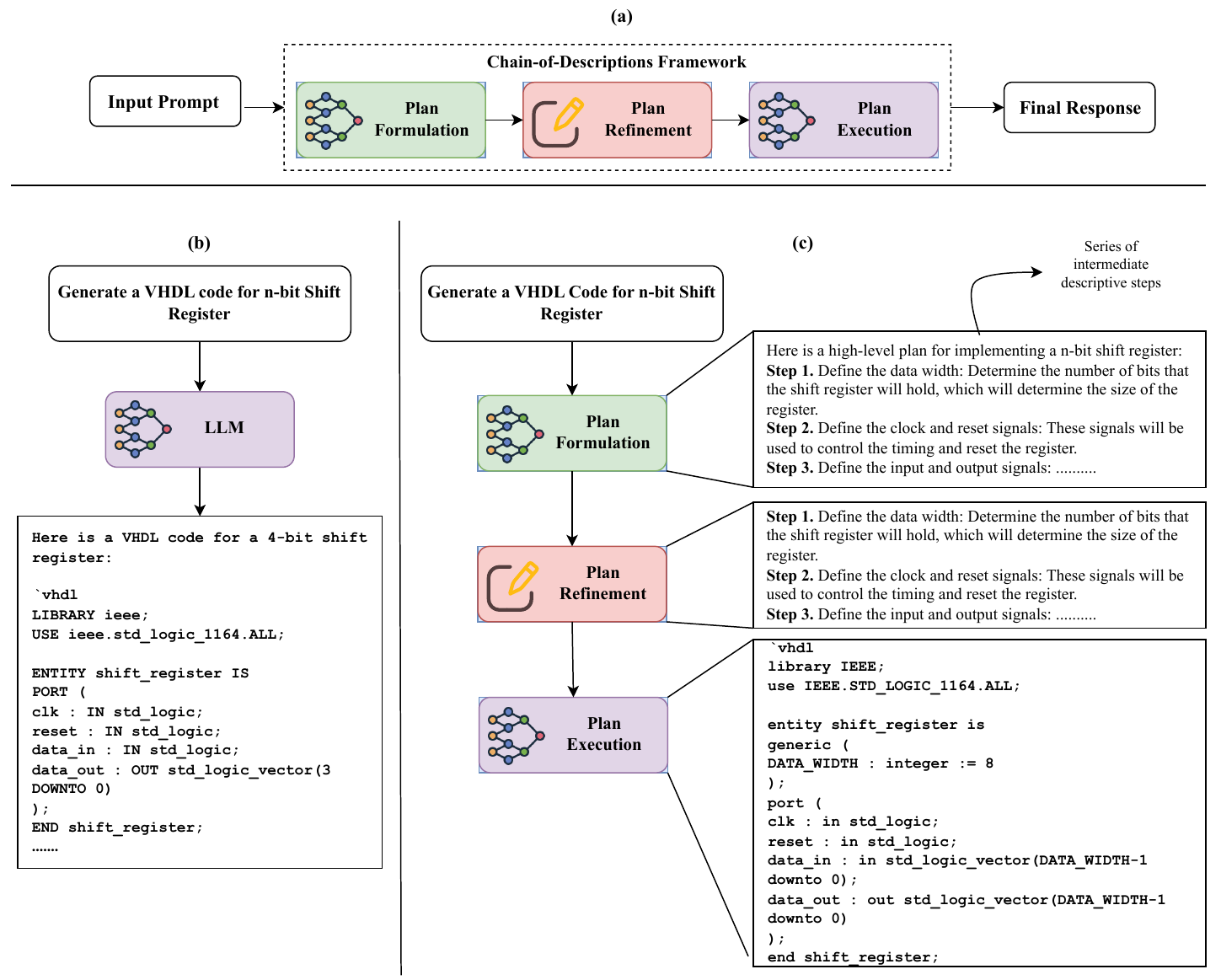}
\caption{(a) Overview of the Chain-of-Description (CoDes) framework: Input prompts and final responses vary between code generation and summarization. (b) Zero-Shot code generation: LLM produces VHDL code for a 4-bit shift register, even when the prompt requests an ``n-bit'' register. (c) CoDes pipeline for an n-bit register: LLM generates a detailed plan with steps that account for the correct data\_width for the requested n-bits.}

    \label{fig:codes_arch}
\end{figure*}
\section{Chain-Of-Descriptions}
The Chain-of-Descriptions (CoDes) framework aims to enhance code LLMs by outlining structured intermediate descriptive steps for VHDL code generation and summarization. Figure \ref{fig:codes_arch} provides both an overview and an example of our CoDes framework. The core concept is that code LLMs, when suitably prompted, can generate and execute a plan as descriptive steps, which are then integrated with the input prompt for an improved response. For VHDL code generation, CoDes first creates a natural language plan that breaks down the task into manageable steps for VHDL synthesis. For VHDL summarization, CoDes provides a step-by-step explanation based on line-by-line analysis. This approach ensures thorough task understanding and completion. The process involves:
(a) \textbf{Formulate a Plan}: Generate intermediate steps using natural language prompts tailored to the task (generation or summarization).
(b) \textbf{Refine the Plan}: Clean and refine LLM responses through a simple post-processing step for clarity.
(c) \textbf{Execute the Plan}: Combine the refined plan with the input prompt to produce the final LLM response.

\subsection{Plan Formulation}
In the Plan Formulation step, the LLM is prompted to generate a series of intermediate descriptive steps, creating a structured plan that guides the model to effectively complete code generation and summarization tasks. This is achieved through natural language prompts tailored to the specific task.

\subsubsection{VHDL Code Generation}
The process begins with a problem statement ($<prob\_st>$), such as ``Design a 4-bit binary counter''. The LLM is then given a natural language prompt that follows a template: ``Generate a plan specifying the intermediate steps to $<prob\_st>$''. An example prompt would be ``Generate a plan specifying the intermediate steps to design a 4-bit binary counter''. This guides the LLM to break down the task into smaller, manageable steps, ensuring a systematic approach to VHDL code synthesis.

\subsubsection{VHDL Code Summarization}
\label{sec:summ}
For the summarization task, the process starts with a given VHDL code snippet. The LLM is prompted to provide a step-by-step explanation, considering the preceding code context. Each prompt follows a template: ``$<code\_context>$ \textbackslash n Explain the following line of code given the context above: \textbackslash n $<code\_line>$'', where $<code\_line>$ is the current line, and $<code\_context>$ includes all prior lines. This method generates detailed descriptive plans. To handle larger code sizes, we also explore an alternative approach using the hierarchical structure and relationships between VHDL elements derived from Abstract Syntax Trees (ASTs). More details are in Appendix \ref{sec:app_ast_cs}, and the impact of these methods is discussed in Section \ref{expt:ast}.


\subsection{Plan Refinement}

After formulating the initial plan, a post-processing step refines LLM responses by removing unnecessary boilerplate text, such as "Answer: Here is a step-by-step approach.." or "Here is the implementation in VHDL..". This ensures clarity and avoids conflicting instructions when combining the plan with the input prompt. The refinement includes: (a) \textbf{Truncating boilerplate text} with regex for conciseness, and (b) \textbf{Ensuring consistency} with a uniform style across steps. Figure \ref{fig:codes_arch}(c) shows an example of the refined plan. More details are in Appendix \ref{sec:prefine}.

\subsection{Plan Execution}


In the final step, the refined plan is integrated with the input prompt to generate the LLM's response. Although techniques like retrieval-augmentation could enhance this process, this work focuses solely on using the LLM for all steps in the CoDes framework. The model handles both planning and strategizing for code generation and summarization. We explore two execution variants: Single-Step and Multi-Step, to assess if breaking down the process offers an advantage over a combined approach.

\subsubsection{Single-Step Execution}

In the Single-Step method, planning and execution occur in a single LLM prompt, instructing the model to outline steps and generate the final output in one interaction. For example: ``Generate a plan to design a 4-bit binary counter, then provide the final VHDL code''. This approach consolidates planning and execution, potentially simplifying the workflow and reducing execution time.

\subsubsection{Multi-Step Execution}

In the Multi-Step method, planning, refinement, and execution are divided into separate LLM prompts. First, the model generates a detailed plan with intermediate steps. This plan is then refined, and a subsequent prompt combines it with the original input to produce the final response. This approach allows for more granular control, ensuring each phase is carefully considered and executed independently.


\begin{table}[]
\small
\caption{Evaluation Results: CoDes for code generation and summarization on VHDL-Eval and VHDL-Xform dataset. \label{tab:codes_eval}}
\begin{tabular}{lcccccc}
\hline
\multicolumn{1}{c|}{\textbf{Models}}  & \multicolumn{4}{c|}{\textbf{VHDL-Eval}}                                                                                                                                                                                                                              & \multicolumn{2}{c}{\textbf{VHDL-Xform}}                            \\ \hline
\multicolumn{1}{l|}{}                 & \begin{tabular}[c]{@{}c@{}}Pass@1\\ (TB)\end{tabular} & \begin{tabular}[c]{@{}c@{}}Pass@1\\ (SEC)\end{tabular} & \multicolumn{1}{c|}{\begin{tabular}[c]{@{}c@{}}SC$_1$\\ (\%)\end{tabular}} & \multicolumn{1}{c|}{\begin{tabular}[c]{@{}c@{}}PR\\ (\%)\end{tabular}} & R$_L$         & \begin{tabular}[c]{@{}c@{}}PR \\ (\%)\end{tabular} \\ \hline
\multicolumn{7}{c}{\textbf{Code Models}}                                                                                                                                                                                                                                                                                                                                          \\ \hline
\multicolumn{1}{l|}{CodeLlama-34b}    & 0.210                                                 & 0.202                                                  & 22.6                                                                       & \multicolumn{1}{c|}{36.6}                                              & 39.2          & 39.5                                               \\
\multicolumn{1}{l|}{Granite-Code-34b} & \textbf{0.254}                                        & \textbf{0.246}                                         & \textbf{26.5}                                                              & \multicolumn{1}{c|}{\textbf{39.0}}                                     & \textbf{40.6} & \textbf{40.0}                                      \\
\multicolumn{1}{l|}{Deepseek-33b}     & 0.189                                                 & 0.169                                                  & 16.1                                                                       & \multicolumn{1}{c|}{27.1}                                              & 36.1          & 29.0                                               \\
\multicolumn{1}{l|}{Granite-Code-20b} & 0.126                                                 & 0.109                                                  & 18.4                                                                       & \multicolumn{1}{c|}{30.5}                                              & 33.9          & 30.9                                               \\ \hline
\multicolumn{7}{c}{\textbf{Instruct Models}}                                                                                                                                                                                                                                                                                                                                      \\ \hline
\multicolumn{1}{l|}{Mixtral-8x7b}     & 0.082                                                 & 0.068                                                  & 11.8                                                                       & \multicolumn{1}{l|}{15.7}                                              & 21.2          & 15.7                                               \\
\multicolumn{1}{l|}{Mistral-7b}       & 0.091                                                 & 0.088                                                  & 6.7                                                                        & \multicolumn{1}{l|}{5.8}                                               & 20.4          & 8.6                                                \\ \hline
\multicolumn{7}{c}{\textbf{Chat Models}}                                                                                                                                                                                                                                                                                                                                          \\ \hline
\multicolumn{1}{l|}{Granite-Chat-13b} & 0.095                                                 & 0.087                                                  & 12.9                                                                       & \multicolumn{1}{l|}{24.3}                                              & 30.2          & 20.9                                               \\
\multicolumn{1}{l|}{Llama-2-Chat-13b} & 0.079                                                 & 0.065                                                  & 10.8                                                                       & \multicolumn{1}{l|}{21.9}                                              & 28.1          & 21.9                                               \\ \hline
\end{tabular}
\end{table}

\section{Experiments}
\label{expt:all}
\begin{figure*}[!ht]
    \centering
    \includegraphics[width=0.95\textwidth]{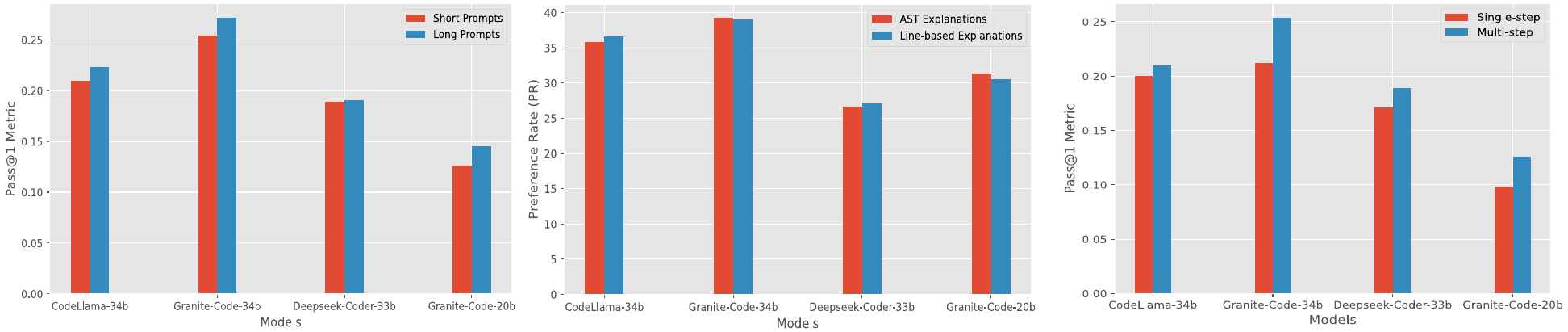}
\caption{Results of Ablation Studies: \textbf{Left:} Short vs. Long Descriptive prompts for plan formulation for Code Generation Task on the VHDL-Eval Dataset using the Pass@1 Metric. \textbf{Center:} AST vs. Line-based explanations for Code Summarization Task on the VHDL-Eval Dataset using the LLM Preference Rate (PR) Metric. \textbf{Right:} Single-Step vs. Multi-Step Plan Execution Strategies for Code Generation Task on the VHDL-Eval Dataset using the Pass@1 Metric. 
}
    \label{fig:ablations}
\end{figure*}

We outline our study designed to address the following research questions: \textbf{(RQ1)} \textbf{Effectiveness of \textsc{CoDes}}: How well does the CoDes framework enhance LLM performance in code generation and summarization? \textbf{(RQ2)} \textbf{Influence of Descriptive Prompts}: Does a longer, detailed problem description improve plan formulation and code generation? \textbf{(RQ3)} \textbf{Impact of Planning Strategies}: How do different planning strategies (AST vs. line-based) affect code summarization? \textbf{(RQ4)} \textbf{Effect of Execution Strategies}: How do single vs. multi-step execution strategies impact LLM performance in code generation and summarization?


  

\subsection{Effectiveness of \textsc{CoDes}}

Table \ref{tab:codes_eval} shows the results of using the CoDes framework with various LLMs for VHDL code generation and summarization, evaluated across different metrics. These results, obtained with the Multi-step plan execution strategy, demonstrate significant performance improvements. The introduction of intermediate descriptive steps enhances the models' ability to generate and summarize VHDL code. Notably, Granite-based models show a $\sim40\%$ improvement in the Pass@1 metric using testbenches compared to Table \ref{tab:zs_eval}. The Granite-Code-34b model consistently outperforms others across metrics for both tasks. The CoDes framework is particularly effective for models that can break down tasks into intermediate steps, validating its efficacy in improving performance on VHDL tasks.

\subsection{Influence of descriptive prompts}

We investigated if detailed problem descriptions enhance code generation. The Pass@1 metric for VHDL code was measured with short and long descriptions. Figure \ref{fig:ablations} (Left) shows that longer descriptions improve performance across all models, supporting our CoDes framework's approach. Detailed prompts help generate better VHDL code.


\subsection{Impact of Planning Strategies}
\label{expt:ast}
In the context of code summarization, we explore the effect of different prompting strategies for plan formulation. Specifically, we leveraged the hierarchical structure and VHDL elements provided by the Abstract Syntax Tree (AST) to prompt the LLM as explained in Appendix \ref{sec:app_ast_cs}. This approach is less time-consuming in comparison to the line-by-line approach discussed earlier in Section \ref{sec:summ}.  Our findings, illustrated in Figure \ref{fig:ablations} (Center), demonstrate that both strategies perform comparably, with marginal improvements in specific evaluation conditions. No single strategy consistently outperforms the other across different models. Future research will explore various hyperparameter settings and methods to better utilize ASTs for code summarization.

\subsection{Effect of Different Plan Execution Strategies}

We examine how Single-step and Multi-step execution variants affect LLM performance in code generation and summarization tasks. Figure \ref{fig:ablations} (Right) shows our code generation evaluation results using both strategies. Multi-step execution significantly improves performance compared to Single-step execution, allowing the model to process and refine intermediate outputs for better final results. In code summarization, as shown in Figure \ref{fig:app_multi_summ} in the Appendix \ref{sec:app_multi}, Multi-step execution consistently outperforms the Single-step approach. Although Single-step offers slight improvements over zero-shot evaluation, Multi-step demonstrates markedly higher performance. Future work will focus on optimizing these strategies and addressing the limitations outlined in Appendix \ref{sec:app_limit} to further enhance model performance.

\section{Conclusion}
In this study, we explore and enhance Large Language Models (LLMs) in the Electronic Design Automation (EDA) domain, focusing on VHDL code generation and summarization tasks. Despite the success of LLMs in natural language processing, their performance in handling hardware description languages (HDLs) like VHDL has been underexplored and found lacking. Our zero-shot evaluation using an augmented dataset reveals significant shortcomings in existing models, highlighting the need for improved methodologies. To address this, we introduced the Chain-of-Descriptions (CoDes) framework, which incorporates intermediate descriptive steps to improve LLM performance for VHDL tasks. Experiments on various LLMs using the VHDL-Eval dataset demonstrate that CoDes markedly enhances VHDL code generation and summarization. Intermediate steps help models better understand and execute tasks, resulting in more coherent outputs. Our findings reveal that longer descriptive prompts significantly enhance code generation performance, while adopting Multi-step execution consistently outperforms Single-step execution for both tasks. Thus, the CoDes framework provides a structured methodology applicable to domains requiring detailed and systematic problem-solving approaches. Through these efforts, we aim to advance the capabilities of code LLMs, making them more effective tools in electronic design automation and beyond.

\bibliographystyle{ACM-Reference-Format}
  \bibliography{base}

\appendix

\section{VHDL-Eval Dataset}
\label{sec:app_vhdleval}
VHDL-Eval benchmark is designed for evaluating VHDL code generation models. This dataset comprises 202 code problems tailored for VHDL. It is constructed by translating problems from the Verilog-Eval dataset \cite{liu2023verilogeval}, which originally consisted of evaluation problems for Verilog, into VHDL. Additionally, publicly available VHDL problem sets were scraped from tutorials to enrich the dataset.

\begin{itemize}
    \item \textit{task\_id}: Indicates the source followed by a problem-specific unique identifier. For example, verilog\_eval/always\_case refers to a problem from the Verilog-Eval dataset, while tutorial/ripple\_carry\_adder denotes a problem scraped from a tutorial.
    \item \textit{declaration}: Function/entity declaration including necessary libraries or packages.
    \item \textit{problem\_statement}: Brief description of the problem.
    \item \textit{short\_description}: A succinct short explanation of the core functionality of the code.
\item \textit{long\_description}: Detailed explanation specifying the functionality along with example input/output.
    \item \textit{prompt}: Commented problem statement concatenated with function/entity declaration.
    \item \textit{canonical\_solution}: Verified solution to the problem.
    \item \textit{testbench}: Test program including test cases.
\end{itemize}

To ensure the model can make inferences and to prevent benchmark leakage, problem statements are reframed to avoid direct reuse. Basic logic gate implementations are modified to involve n-bit implementations or altered to be part of a larger problem. For example, ``Create a VHDL code for computing a logical OR of all the bits in the input'' is changed to ``Assess whether the sum of all input bits is greater than zero. Output 1 if the sum is greater than zero; otherwise, output 0''.

\section{VHDL-Xform Dataset}
\label{sec:app_clone}
\begin{figure*}[!ht]
    \centering
    \includegraphics[width=\textwidth]{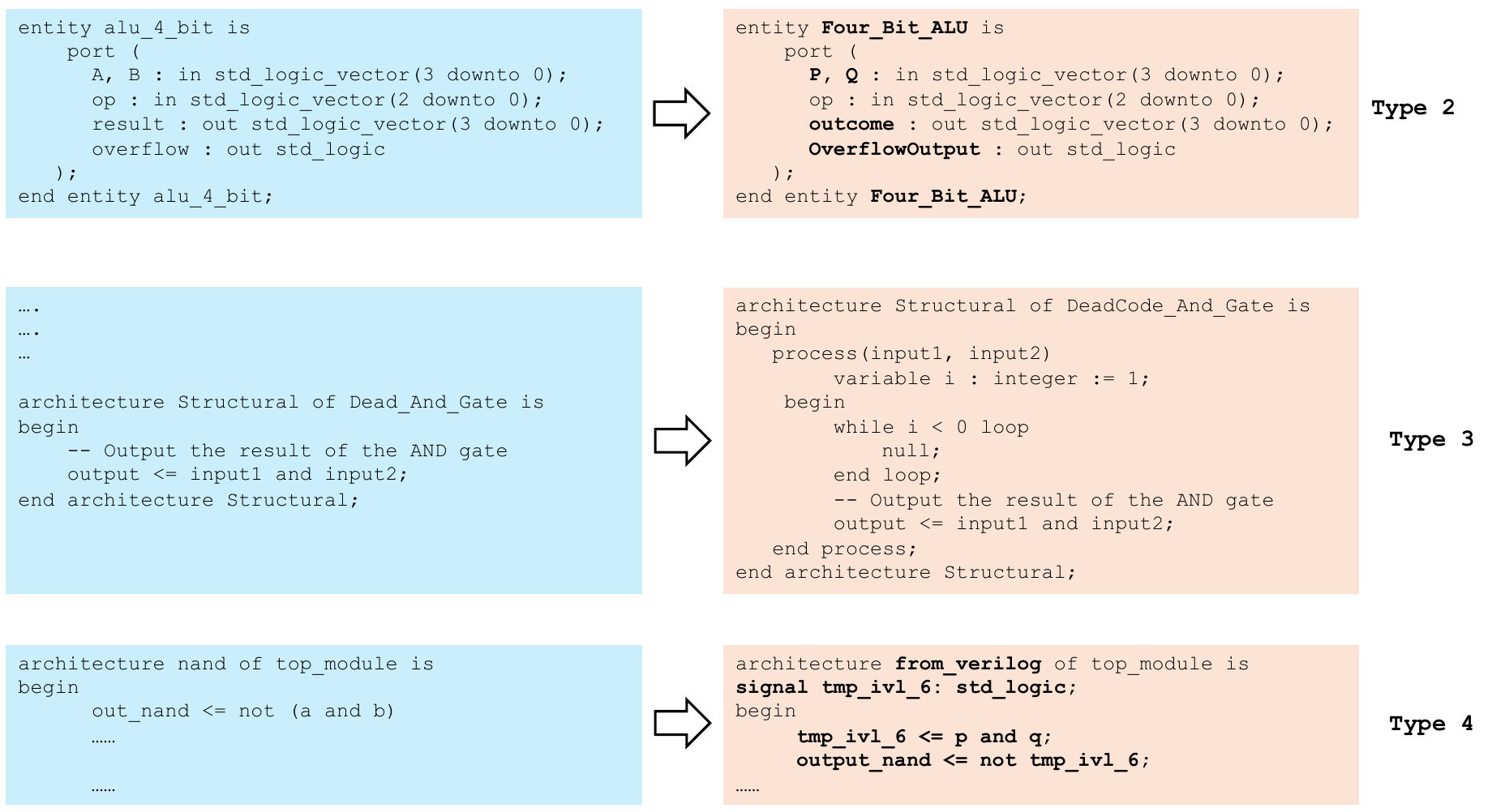}
\caption{Samples of different transformation strategies applied using the three categories of code clones -- Type 2, Type 3 and Type 4.}
    \label{fig:clone_types}
\end{figure*}

Figure \ref{fig:clone_types} displays samples from the VHDL-Xform dataset, illustrating the transformation strategies applied based on the different categories of code clones.  Table \ref{tab:xform} presents the dataset statistics of our VHDL-Xform dataset.

\section{LLM Baselines}
\label{sec:bl}
In this section, we provide more details on the different LLM baselines used in this work:
\subsubsection{Code Models}
\begin{itemize}
    \item \textbf{CodeLlama-34b}: Code Llama\footnote{\url{https://huggingface.co/codellama}} is an instruction-tuned, 34-billion parameter auto-regressive language model that uses an optimized transformer architecture for general code synthesis and understanding.
    
    \item \textbf{Granite-Code-34b}: Granite-34B-Code-Instruct\footnote{\url{https://huggingface.co/ibm-granite/granite-34b-code-instruct}} is a 34-billion parameter model fine-tuned from Granite-34B-Code-Base on a combination of permissively licensed instruction data to enhance instruction-following capabilities, including logical reasoning and problem-solving skills.
    
    \item \textbf{Deepseek-Coder-33b}: DeepSeek Coder\footnote{\url{https://github.com/deepseek-ai/DeepSeek-Coder}} is composed of a series of code language models, each trained from scratch on 2T tokens, with a composition of 87\% code and 13\% natural language in both English and Chinese. Specifically, we use the 33-billion parameter model for our evaluation.
    
    \item \textbf{Granite-Code-20b}: Granite-20B-Code-Instruct\footnote{\url{https://huggingface.co/ibm-granite/granite-20b-code-instruct}} is a 20-billion parameter model fine-tuned from Granite-20B-Code-Base to enhance instruction-following capabilities similar to the 34-billion parameter models.
\end{itemize}

\subsection{NL Instruct Models}
\begin{itemize}
    \item \textbf{Mixtral-8x7b}: The Mixtral-8x7B\footnote{\url{https://huggingface.co/mistralai/Mixtral-8x7B-Instruct-v0.1}} Large Language Model (LLM) is a pretrained generative Sparse Mixture of Experts. The Mixtral-8x7B is known to outperform Llama-2-70B on many benchmarks.
    
    \item \textbf{Mistral-7b}: The Mistral-7B-Instruct-v0.2\footnote{\url{https://huggingface.co/mistralai/Mistral-7B-Instruct-v0.2}} Large Language Model (LLM) is an instruct fine-tuned version with a 32k context window.
\end{itemize}
\begin{table}[]
\centering
\caption{Dataset Statistics of VHDL-Xform dataset}
\label{tab:xform}
\begin{tabular}{@{}lr@{}}
\toprule
\multicolumn{2}{c}{\textbf{\begin{tabular}[c]{@{}c@{}}VHDL-Xform \\ Dataset Statistics\end{tabular}}} \\ \midrule
\# Total VHDL Code-Clone Pairs                                 & 6,500                                 \\
\% of Type-2 Clones                                           & 30.2\%                                \\
\% of Type-2 Clones                                           & 39.4\%                                \\
\% of Type-4 Clones                                           & 30.4\%                                \\ \bottomrule
\end{tabular}

\end{table}

\begin{figure*}[!ht]
    \centering
    \includegraphics[width=0.8\textwidth]{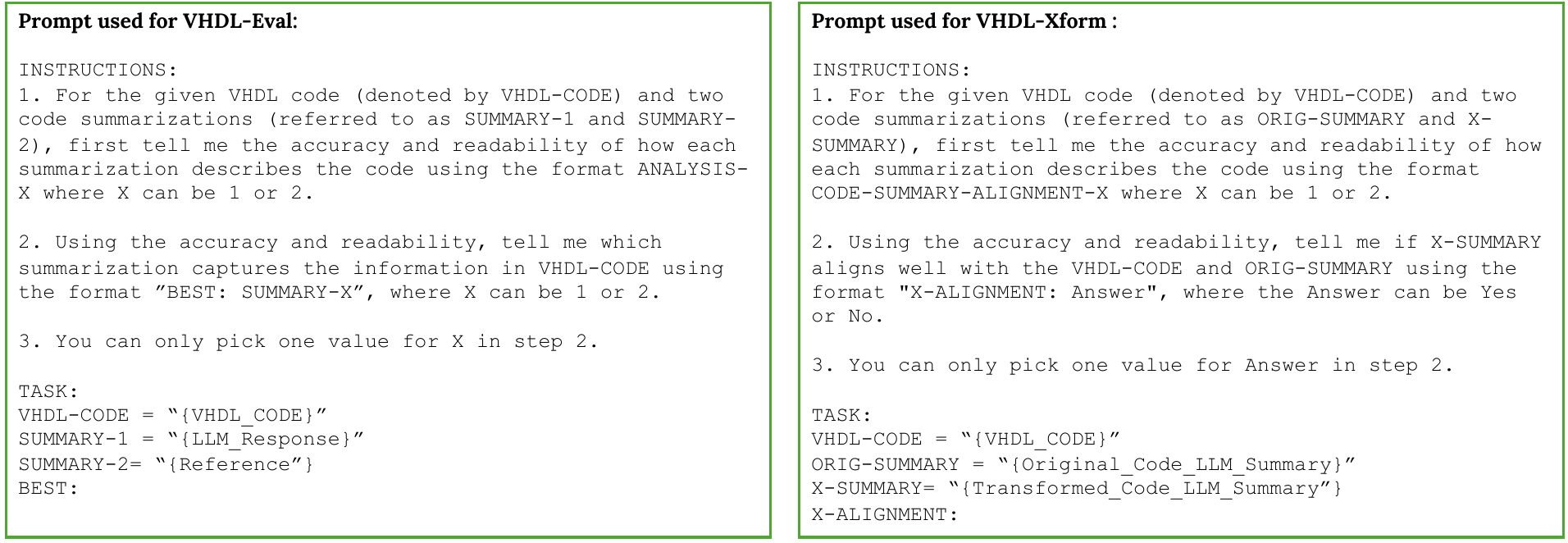}
\caption{Prompt used for LLM judgment -- \textbf{Left}: Prompt for computing the preference rate for VHDL-Eval dataset. \textbf{Right}: Prompt for computing the preference rate for VHDL-Xform dataset.}
    \label{fig:prompt_llr}
\end{figure*}

\subsection{Chat Models}
\begin{itemize}
    \item \textbf{Granite-Chat-13b}: Granite Chat is a 13-billion parameter version of IBM models optimized for dialogue use cases.
    
    \item \textbf{Llama-2-Chat-13b}: Llama-2 Chat\footnote{\url{https://huggingface.co/meta-llama/Llama-2-13b-chat}} is the 13-billion parameter version of pretrained and fine-tuned generative text models from Meta optimized for dialogue use cases.
\end{itemize}

\section{Evaluation Settings}
\label{sec:eval_settings}
Drawing from previous research \cite{wang2022self,min2023beyond}, we conducted experiments within defined evaluation parameters. These included temperature sampling with $\mathcal{T} = {0.5, 0.7}$ and truncation at the top-$k$ tokens ($k = 40$) with the highest probability. To mitigate redundant text generation, a repetition penalty of 1.2 was applied. In the zero-shot scenario, we primarily present scores derived from single sampling from the LLMs. Additionally, we explored using a top-$k$ value set to 50 and a temperature of $\mathcal{T} = 0.89$ for specific LLMs during the plan formulation phase.

\section{LLM Preference Rate}
\label{sec:pr_more}
Figure \ref{fig:prompt_llr} displays the prompts used to compute the LLM Preference Rate (PR) explained in Section \ref{sec:pr}.

\subsection{Plan Refinement}
\label{sec:prefine}
The plan refinement phase extracts intermediate descriptive steps using a set of rules: (a) identifying text separated by a colon ``:'' followed by a new line, (b) detecting explicit patterns such as ``Step \textbackslash d+'' or ``\textbackslash d+\textbackslash .'' representing each step as ``Step 1'' or ``1.'', and (c) splitting the text by new lines if the previous patterns are not found. If the generated plan does not contain any of these patterns, we sample the plan up to three times to extract intermediate steps. If no intermediate steps are found, we return an empty string to avoid incorporating noisy or conflicting information into the plan.



\begin{table*}[h]
\centering
\caption{Template Prompt Examples for AST Nodes}
\label{tab:template_prompts}
\begin{tabular}{l|l}
\hline
\textbf{AST Node}      & \textbf{Template Prompt Example}                                              \\ \hline
Entity                 & Explain the purpose of the entity named ``{entity\_name}'' along with its port information.                             \\ 
Architecture           & Explain the functionality and behavior implemented in the architecture named ``{architecture\_name}'' for the entity ``{entity\_name}''.                             \\ 
Process                & Explain the behavior within the process named ``{process\_name}''.     \\ 
Components     & Explain the role and instantiation details of the component ``{component\_name}'' in the architecture ``{architecture\_name}''.                       \\ 
Procedure & Explain the functionality of the procedure named ``{procedure\_name}'' and its intended functionality.       \\ 
Function  & Explain the purpose and return value of the function named ``{function\_name}''.      \\ \hline
\end{tabular}

\end{table*}

\section{AST for Code Summarization}
\label{sec:app_ast_cs}
Given a VHDL code, an Abstract Syntax Tree (AST) captures the hierarchical tree structure and the relationships between various elements such as entities, architectures, signals, processes, and control statements. Each node of the tree denotes a construct occurring in the VHDL code. By examining the nodes and their relationships, we can extract key aspects of the code's functionality and structure, which can then be transformed into concise and informative summaries. The AST nodes categorize the constructs, allowing us to identify and summarize different parts of the VHDL code systematically. The potential for code summarization using AST lies in its ability to represent the syntactic structure of VHDL code in a detailed manner. By traversing the AST, we can create template prompts that capture the essence of each structural element. For example, for an entity declaration, we use a prompt that provides the VHDL code and ask the model to summarizae the purpose and ports of the entity. Similarly, for an architecture block, we can describe its functionality and behavior. This structured approach ensures that each aspect of the code is adequately summarized, providing a comprehensive overview. Table \ref{tab:template_prompts} demonstrates example template prompts used for different non-leaf AST nodes.

In this paper, we briefly explore the potential of AST-based prompts for VHDL code summarization, which opens avenues for more sophisticated and detailed code analysis. By investigating various other prompts and combinations of AST nodes, future research can enhance the precision and depth of code summaries. For instance, combining information from multiple nodes or considering higher-level constructs like configurations and packages could provide richer summaries. This approach not only aids in understanding and maintaining VHDL code but also sets a foundation for automated documentation and intelligent code assistance tools.

\section{Effect of Single vs. Multi-Step Execution}
\label{sec:app_multi}
\begin{figure}[!ht]
    \centering
    \includegraphics[width=\linewidth]{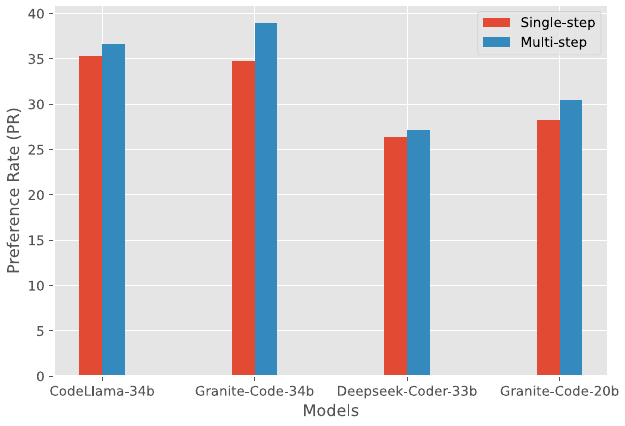}
\caption{Single-Step vs. Multi-Step Plan Execution Strategies for Code Summarization Task on the VHDL-Eval Dataset using the Preference Rate (PR) Metric}
    \label{fig:app_multi_summ}
\end{figure}

Figure \ref{fig:app_multi_summ} illustrates the evaluation outcomes of using single and multi-step execution strategies for VHDL code summarization. Like in the VHDL code generation task, we find that the multi-step approach performs better than the single-step approach for code summarization as well.

\section{Limitations}

\label{sec:app_limit}
In our study, we introduce CoDes, or chain-of-description, as a novel approach to augmenting LLMs for code summarization and generation tasks. This method heavily relies on the quality of intermediate descriptive steps provided by the LLM being utilized. Specifically, during the code generation phase, we implement a strategy of sampling up to three times to verify if the intermediate steps are generated according to the expected pattern for extraction. However, when it comes to code summarization, while a line-by-line description may contribute to successful code generation, it may not always serve as an adequate explanation for human comprehension. This limitation arises from the inherent difficulty of such descriptions in conveying a higher-level understanding of a piece of code's functionality. Consequently, the quality of intermediate steps significantly impacts the overall performance of the model for both tasks.

Furthermore, the datasets used in our study feature self-contained problem sets and solutions, lacking the complexities found in real-life design scenarios. While this simplification facilitates the generation of intermediate descriptive steps by LLMs, it poses challenges for models when confronted with non-straightforward problems and increased complexity. Similarly, in the context of code summarization, employing a line-by-line approach becomes increasingly time-consuming as the number of lines of code increases. To address these limitations, future research endeavors could focus on evaluating CoDes for more complex RTL designs while simultaneously exploring methods to reduce processing time without sacrificing performance. Moreover, it is essential to recognize that the evaluations conducted in our study represent preliminary explorations of openly accessible models. While other prominent and powerful models could serve as judge LLMs, we opted to utilize the Llama-3-70B model due to its open-source nature and accessibility. Although this model may exhibit some inaccuracies, it remains a potent tool that has outperformed numerous open-source chat models on industry benchmarks. However, future research could delve deeper into exploring the variation in evaluation scores by leveraging different judge LLMs. Additionally, conducting further experiments on a wider array of code LLMs, exploring various hyperparameters, holds promise for yielding improved results and insights, a task we defer to future investigation.

\end{document}